\documentclass[10pt,twocolumn,letterpaper]{article}

\usepackage{cvpr}
\usepackage{times}
\usepackage{epsfig}
\usepackage{graphicx}
\usepackage{amsmath}
\usepackage{amssymb}
\DeclareMathOperator*{\argmax}{argmax}
\usepackage{algorithm}
\usepackage{algpseudocode}
\usepackage{color}
\usepackage{graphicx}
\usepackage{subfigure}

\cvprfinalcopy 

\def\cvprPaperID{} 

\ifcvprfinal\fi
\begin{document}

\title{Graph Neural Based End-to-end Data Association Framework for Online Multiple-Object Tracking}

\author{Xiaolong Jiang $^{*}$\\
{\tt Beihang University}
\and
Peizhao Li \thanks{Equal Contribution}\\
{\tt\ Brandeis University}
\and
Yanjing Li\\
{\tt\ Beihang University}
\and
Xiantong Zhen\\
{\tt\ Inception Institute of Artificial Intelligence}
}

\maketitle
\thispagestyle{empty}

\begin{abstract}
In this work, we present an end-to-end framework to settle data association in online Multiple-Object Tracking (MOT). Given detection responses, we formulate the frame-by-frame data association as Maximum Weighted Bipartite Matching problem, whose solution is learned using a neural network. The network incorporates an affinity learning module, wherein both appearance and motion cues are investigated to encode object feature representation and compute pairwise affinities. Employing the computed affinities as edge weights, the following matching problem on a bipartite graph is resolved by the optimization module, which leverages a graph neural network to adapt with the varying cardinalities of the association problem and solve the combinatorial hardness with favorable scalability and compatibility. To facilitate effective training of the proposed tracking network, we design a multi-level matrix loss in conjunction with the assembled supervision methodology. Being trained end-to-end, all modules in the tracker can co-adapt and co-operate collaboratively, resulting in improved model adaptiveness and less parameter-tuning efforts. Experiment results on the MOT benchmarks demonstrate the efficacy of the proposed approach.
\end{abstract}

\section{Introduction}
Given a video sequence, Multi-Object Tracking (MOT) algorithms generate consistent trajectories by localizing and identifying multiple targets in consecutive frames. Considering its spatial-temporal nature, MOT task is intrinsically complicated for claiming a formidable solution search space. Moreover, the complications of MOT further aggravates with the increasing number of targets, complex object behaviors, and intricate real-life tracking environments.

Aiming at decoupling the combinatorial complications, most trackers solve the object localization and identification separately and lead to two categories of MOT algorithms. On one hand, the Tracking-by-Prediction methods \cite{TbyP1, TbyP2, CyberneticsMine} prioritize object identification by deploying multiple Single Object Trackers (SOTs) on the basis of motion prediction. However, due to the absence of detections, these methods are troubled to adapt to the varying object number because of the object birth \& death (i.e. object entering or leaving the scene). On the other hand, the Tracking-by-Detection methods first localize objects anonymously with detectors, then resolve object identification via data association \cite{OnlineMOTCVPR2014, MulticutTang2017, Cvpr18TwoFoldSiamese}.

In this work, we follow the Tracking-by-Detection strategy. Taking detections as a given, the core of our proposed tracker is its data association module. To achieve online tracking capability, the tracker performs frame-by-frame data associations which can be graphically formulated as Maximum Weighted Bipartite Matching problems. For each pair of consecutive frames, a weighted bipartite graph is constructed involving trajectories in the previous frame and detection responses in the current frame. The matching problem established whereupon is resolved by first generating pairwise affinities as edge weights, then solving the obtained optimization problem to generate the association output.

Accordingly, the data association module starts with generating pairwise affinities. In tracking scenes baffled with target appearance variations and similar distractors, the expressivity and discriminability of the computed affinities is determined by the adopted feature representation method, as well as the distance metric deployed to quantify the affinities. Earlier approaches leverage advanced hand-crafted features \cite{HandAppr1, HandAppr2, HandAppr3} to achieve robust representation. More recently, CNN based deep features are widely exploited \cite{MulticutTang2017, Eccv18BilinaerLSTM, CVPR2017Quadruplet,  PAMI2018DeepApprMOT} instead. Furthermore, a multi-cues strategy has also been practiced to supplement the appearance cue with others \cite{LearnedApprCVPR2010, HandAppr2, TwoStream1}, among which motion is the most vastly adopted \cite{TwoStream2, TwoStream3, TwoStream4, TrackingTheUntrackable, jiang2019model}. On the basis of the encoded feature vectors, hand-engineered distance measures \cite{HandAppr1, BhattacharyyaDist1, BhattacharyyaDist2} are generally utilized to compute the affinity scores. In addition, attempts have also been made to learn metrics that can co-adapt with the feature learning altogether \cite{MWIS2011, TrackingTheUntrackable, Icip2018MOTSiameseLSTM}.

Given the computed affinities, the following optimization problem defined on the weighted bipartite graph is normally configured into a linear assignment formalism and solved with well-designed optimizers or heuristics \cite{BM1WACV2014, BM2ShahCVPR2012}. However, these approaches suffer from tedious designing efforts, prohibitive computation expense, and poor scalability. Particularly, in the presence of frequent object birth \& death, the combinatorial formalism of linear assignment constraints are violated, thus inducing erroneous optimization results which in turn leads to false associations. As a solution, we strive to resolve the optimization problem relying on the function approximation capacity of deep networks in a data-driven way. Nevertheless, this approach is non-trivial to realize. Firstly, although deep neural networks such as CNN and RNN have exceptional feature learning capability, yet their capacities to conduct relational reasoning for data association are limited; Secondly, the varying number of targets give rise to changing dimensionality of the association problem, demanding the otherwise fixed model to be adaptive; Moreover, available data is limited for tracking problem to support the training of heavy models.

Inspired by its graphical formulation of the optimization problem, we observe Graph Neural Network (GNN) \cite{TheGNNModel} is well-suited to solve the problem. By reasoning over non-Euclidean graph data in a message-passing way, the proposed GNN optimization module is endowed with improved relational reasoning capacity and can cope with the varying cardinality problem via the deployments of localized operations. Furthermore, the module is light-weight and converges well. By integrating the aforementioned affinity learning module end-to-end, all parameters in the data association pipeline can co-adapt and co-operate compactly, results in better model adaptiveness, scalability, and efficiency with acceptable model complexity. For the purpose to better optimize the complicated network with diverse modules, we design the multi-level matrix loss which is assembled to enhance the training performance. The main contributions of this work include:
\begin{itemize}
\item We propose an end-to-end framework incorporating affinity learning and optimization modules to solve the data association problem in online multiple-object tracking.
\item We design the optimization module with Graph Neural Network (GNN), which learns to solve the constructed maximum weighted bipartite matching problem in a data-driven way, avoiding excessive algorithm design and parameter tuning efforts.
\item We employ assembled supervision in conjunction with the proposed multi-level matrix loss to ensure the training performance of the end-to-end network composing diverse modules.
\item We demonstrate experimentally that the GNN optimization module improves data association performance, and overall our method yields competitive results with other state-of-the-art trackers on the MOT benchmark.
\end{itemize}

\begin{figure*}[ht]
\centering
\includegraphics[scale=0.36]{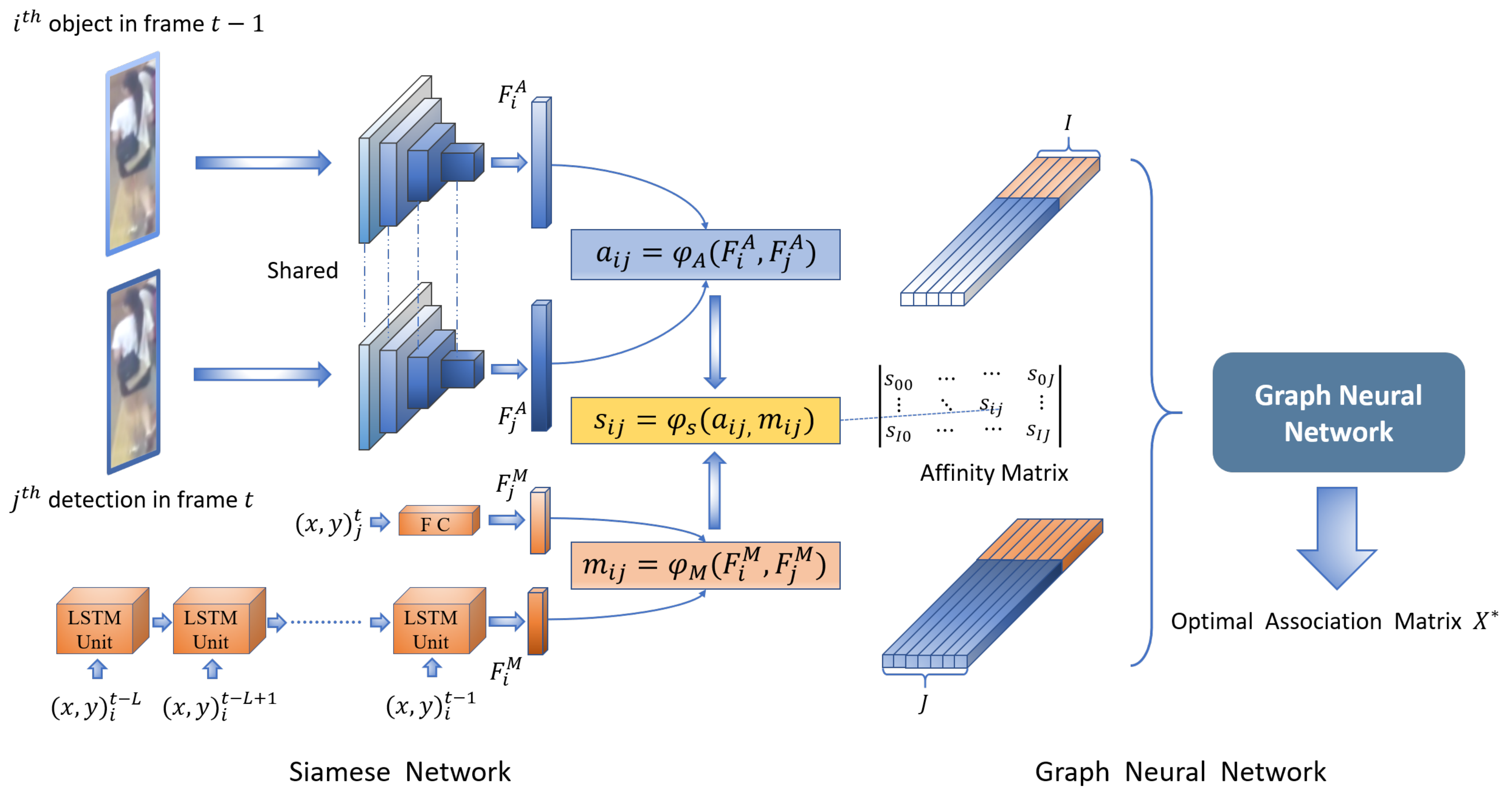}
\caption{The pipeline of the proposed framework. It consists of the Siamese Network for affinity computation and the Graph Neural Network for optimization for the in-complete bipartite graph. The figure is best viewed in color.}
\label{fig: pipeline}
\end{figure*}
\section{Related Work}
\textbf{Affinity Computation in Data Association.} Data association is the process of dividing a set of instances into different groups, such that to maximize the global cross-group similarities while maintaining one-to-one association constraint. This fundamental technique exists in various domains that involve correspondence matching \cite{PhdThesisDA}, such as person re-identification \cite{Cvpr2018MTMC, ReIdSurvey, ReIDDA1}, keypoint matching \cite{DLGraphMatching}, 3D reconstruction \cite{PAMI3DReconShah}, action recognition \cite{DA4ActionRecog}, and T-by-D based MOT \cite{LearningByTracking}.

Affinity computation lays the foundation for data association. It provides the similarity measures upon which the maximization is established. In the context of multi-object tracking, pairwise or tracklets-based appearance affinity scores are computed by first extracting reliable feature representations with hand-crafted features \cite{HandAppr1, HandAppr2, HandAppr3, ColorApprCVPR2015, TextureAppr, BrightnessAppr, ICCV2015LocalFlow}, learnable features \cite{ApprAdaBoost, ApprMotionMIL, EnsembleTracking, LearnedApprCVPR2010, LearnedApprECCV2012, LearningToAssociate}, or deep features \cite{PAMI2018DeepApprMOT, DCNNApprFeatureNIPS2013, DeepTrackTIP2016DeepApprFeature}. For the purpose to supplement the appearance cues under severe variations, attempts have been made to jointly investigate multiple cues \cite{LearnedApprCVPR2010, HandAppr2, TwoStream1}. Amongst, motion information is the most widely applied \cite{ApprMotionMIL, DA1, TwoStream2, TwoStream3, TwoStream4, TrackingTheUntrackable}. The similarity between paired feature vectors is quantified by a distance metric such as Euclidean, Mahbolinios or Bhattacharyya distance \cite{HandMadeDistanceMetric1, MilanEnergy, OnlineMOTCVPR2014, HandAppr1, BhattacharyyaDist1, BhattacharyyaDist2}. Moreover, metric learning has also been deployed in \cite{MWIS2011, TrackingTheUntrackable, Icip2018MOTSiameseLSTM} to learn adaptive metric from data. Noteworthily, end-to-end affinity computation has been proposed with deep Siamese architecture \cite{SiameseLeCun, LearningByTracking, TripletLossTrackingEccv18, Cvpr18TwoFoldSiamese, Cvpr2018RASnet, Cvpr2018MTMC, Icip2018MOTSiameseLSTM}.

\textbf{Optimization in Data Association.} Data association can be interpreted as the Set Partition Problem (SPP)\cite{SPP}. In multi-object tracking, this SPP formalism is specialized into the Multi-Dimensional Assignment (MDA) problem, which describes the optimization procedure defined on a k-partite graph to partition object observations into trajectories cross $k$ frames. Offline methods indicate $k>=3$ such that the data association is executed in a batch mode. Such a global association strategy is robust and accurate, yet inevitably introduce NP-hardness and forfeit the real-time tracking capability. Tracklet-based \cite{TrackletsBasedECCV2008, SOTMOT, ECCV18YangMHDualAttention, SOTMOT2} and detection-based \cite{MilanEnergy, NevatiaCVPR2008} methods have been practiced to realize offline multi-object tracking. Differently, online multi-object trackers \cite{BM1WACV2014, BM2ShahCVPR2012, OnlineMOTCVPR2014} perform frame-by-frame associations with real-time tracking capability recovered. Online methods are usually graphically formulated as the Maximum Weighted Bipartite Matching \cite{DLGraphMatching} problem and solved with the Hungarian algorithm \cite{Hungarian}. A variety of other graphical formalisms have also been devised to settle the data association, including Network Flow \cite{ShahNF,PAMI16NF,NF2CVPR2011, DeepNF}, Minimum Cost Multi-cut \cite{Multicut2015CVPR, MulticutECCV2016, MulticutTang2017}, Maximum-weight Multi-Clique \cite{ShahGMMCP2015CVPR}, etc. On the basis of these optimization formulations, attempts have been made to solve the problem in a data-driven manner \cite{DeepNF, DLGraphMatching}. Particularly, in \cite{DeepTracking, ondruska2}, Ondruska et al. initially propose to deploy recurrent neural networks (RNNs) in solving MOT on a primitive level without formal data associations. Following this line of research, in \cite{Milan2017AAAI} Milan et al. established the first end-to-end online multi-object tracking method with explicit data association realized using RNNs.

\textbf{Graph Neural Network}. Common neural networks models are designed to work with Euclidean data as inputs. Graph neural networks, as neural network models operating on non-Euclidean graph data, refers to a structured architecture to conduct graph-to-graph relational reasoning computations \cite{GNNSurvey, DeepMindGNN}. In general, GNN manages to learn complicated semantic information by building them from the lower level in a hierarchical way \cite{DLGoodFellow}. In actualization, this GNN hierarchy is established by aggregating local information from edges to nodes, then to the global level in a message passing manner \cite{MessagePassing}. Over its development in the past decade \cite{GNNOrigin1,GNNOrigin2,GNNorigin3}, GNN models have found far-reaching applications across domains including supervised \cite{GNNSupervised}, semi-supervised \cite{Kipf2017ICLR}, few-shot \cite{GNNFewShot}, and reinforcement learning \cite{GNNReiforcement}, resulted in fruitful well-established networks such as graph convolutional networks \cite{Kipf2017ICLR, MPNN, GAT}, graph-based generative models \cite{MoIGAN, GRNN}, graph-based adversarial learning \cite{GraphSGAN, PeerNets}, etc. Closely related to the proposed work, a few attempts have been taken to settle the Quadratic Assignment Programming (QAP) problem with GNN. In \cite{GNNQAP1}, the general GNN architecture proposed in \cite{GNNorigin3} is adopted to solve the QAP in the context of graph matching. In \cite{GNNCO}, a unique combination of reinforcement learning and graph embedding is realized to resolve the NP-hard combinatorial optimization.

\section{End-to-end Data Association}
Using affinity measures as edge weights, the data association of online multi-object tracking can take the graphical form of a Maximum Weighted Bipartite Matching problem. This matching problem can be formulated into a linear assignment and solved with well-defined algorithms \cite{daFirst, milanDA}. In this section, we present our end-to-end data association model, wherein the affinity computation and optimization module are jointly trained to achieve co-adaptation and co-operation. More importantly, the varying cardinality of the optimization problem caused by object birth \& death is settled in a data-driven way.

\subsection{Problem Formulation}
For online data association, one bipartite graph $G_{t}$ is constructed between every pair of consecutive frames, where the two disjoint sets each contain the existing trajectories $T^{t-1}=\{{{T}_{1}^{t-1}},{{T}_{2}^{t-1}},\ldots ,{{T}_{i}^{t-1}}\}$ in the previous frame $t-1$, and the newly detected object observations $O^{t}=\{{{o}_{1}^{t}},{{o}_{2}^{t}},\ldots ,{{o}_{j}^{t}}\}$ in the current frame $t$. $i \in I$ and $j \in J$, where $I$ and $J$ defines the cardinality of the association problem. Particularly, trajectory ${T}_{i}^{t-1}$ is represented by its bounding box observation $t_{i}^{t-1}$ (i.e. the (x, y, w, h) annotation) at frame $t-1$ and its short tracklet of coordinates. The graph is weighted by the affinity matrix ${{S}}\in {\mathbb{R}^{I\times J}}$, where affinity score ${{s}_{ij}}\in \mathbb{R}$ is associated as the weight on the edge between trajectory $i$ and observation $j$. Each edge is also associated with a binary indicator $x_{ij}$ in the association matrix $X$. The association result is the solution of a Maximum Weighted Bipartite Matching problem defined on $G_{t}$, i.e. the optimal association matrix $X^*$ of the corresponding linear assignment defined as follows:
\begin{equation}
{{X}^{*}}={\mathop{\argmax}_{X}{S}^{T}}\bullet X
\label{equa:opt1}
\end{equation}
\begin{equation}
s.t.\left\{ \begin{aligned}
 & \forall j:\sum\limits_{i}{{{x}_{ij}}\le 1} \\
 & \forall i:\sum\limits_{j}{{{x}_{ij}}\le 1} \\
 & {{x}_{ij}}\in \{0,1\} \\
 & \sum\limits_{ij}{{{x}_{ij}}=k,k\le \min (I,J)} \\
\end{aligned} \right.
\label{equa:opt2}
\end{equation}
The computation in (\ref{equa:opt1}) denotes dot-product of two vectorized matrices. The first two constraints in (\ref{equa:opt2}) ensure the assignment feasibilities that no two trajectories can claim the same observation at the same frame. The last constraint computes matrix norm of $X$, indicates that there are exactly $k$ one-to-one associations which satisfy the equality constraint. In other words, $k$ one-to-one associations are established across two frames whereas the rest are birth \& death associations. In cases where $I\ne J$, zero nodes, and edges need to be augmented into the graph to maintain symmetry in matrices $S$ and $X$ so that Hungarian algorithm can be applied to solve the assignment.

Nonetheless, the application of the Hungarian algorithm or similar alternatives is not ideal for all tracking scenarios. For one, such algorithms scale poorly with the increasing problem size and easily becomes intractable in real tracking scenes. More seriously, $k$ is not known as a $prior$ during tracking due to irregular object birth \& death, so the last constraint in (\ref{equa:opt2}) does not always hold, i.e. the optimization formulated as above cannot enclose all association scenarios. In these cases, the availability of combinatorial optimization algorithms is invalidated.

Aiming at achieving an efficient optimization well-compatible to real tracking scenes, we establish a module leveraging the GNN to approximate an optimization solution via supervised learning. In addition, an affinity learning module is proposed to compute the $A$ matrices. Jointly, an end-to-end data association module is realized enabling the co-adaptation and co-operation of both modules collaboratively.

\subsection{Affinity Learning Module}
\label{aff}
Affinity is the quantification of similarity between observations, affinity computation is the starting point of a variety of matching-based tasks \cite{ReIdSurvey, DLGraphMatching, LearningByTracking, PAMI3DReconShah, DA4ActionRecog}. In the context of online multi-object tracking, an affinity matrix ${S}\in {\mathbb{R}^{I\times J}}$ is computed to weight the bipartite graph. Each element $s_{ij}$ in $S$ indicates the similarity between trajectory $t_{i}$ and the newly detected observation $o_{j}$. In the proposed tracker, $s_{ij}$ is computed end-to-end from input bounding box annotations to output scalar value affinity scores through the proposed two-stream affinity learning module, where appearance affinity is quantified with a Siamese Convolutional Neural Network, while motion affinity is investigated based on proximity reasoning with an LSTM motion prediction component. A set of fully connected layers are deployed as the learnable metric to integrate the two.

\subsubsection{Appearance Affinity} Appearance cue provides the most defining image evidence to recognize and discriminate an object. In presences of appearance variations and similar distractors, robust feature representation is vital to achieve reliable affinity. In our method, appearance affinity is computed with a Siamese CNN feature encoding architecture. As shown in Figure \ref{fig: pipeline}, $p_{i}$ and $p_{j}$ are encoded into feature vectors $F_{i}^{A}$ and $F_{j}^{A}$ with dimension of $D_{A}$. The appearance affinity score $s_{ij}$ is computed using $F_{i}^{A}$ and $F_{j}^{A}$ via the learnable metric described later in this section. Differ from the early fusion strategy adopted in \cite{LearningByTracking} where pairs of instances are stacked before input to the Siamese network, we opt to fuse $F_{i}^{A}$ and $F_{j}^{A}$ later as both of them are needed in the GNN optimization module.

\subsubsection{Motion Affinity} Motion cue offers generic and appearance-invariant information to characterize a object according to its dynamic behavior. It has been proven to be a beneficial complement to reinforce appearance cue in cases of severe appearance variations and cluttered background \cite{TwoStream2, TwoStream3, TrackingTheUntrackable}. As shown in Figure \ref{fig: pipeline}, motion affinity is computed on the basis of motion prediction with a LSTM motion model. For a trajectory $T_{i}^{t-1}$ in the previous frame, we maintain a short tracklet $Tr_{i}^{t-1}$ of $T_{i}^{t-1}$ as a sequence of 4D bounding box annotations $Tr_{i}^{t-1} = [{{{t}_{i}^{t-1}},{{t}_{i}^{t-2}},..., {{t}_{i}^{t-L}}}]$, where $L$ denotes the length of the tracklet. Accordingly, the LSTM network unrolled into $L$ time steps with hidden unit size as $D_{M}$. By feeding in a tracklet $Tr_{i}^{t}$, the LSTM generate motion feature vector $F_{i}^{M} \in \mathbb{R} {^{{D}_{M}}}$. On the other hand, the bounding box annotation of a observation from the current frame $o_{j}$ is encoded by a set of fully-connected layers into feature vector $F_{j}^{M} \in \mathbb{R} {^{{D}_{M}}}$. The motion affinity is then computed based on $F_{i}^{M}$ and $F_{j}^{M}$ utilizing the learnable distance metric.

\subsubsection{Metric Learning} Aiming at end-to-end module adaptation and cooperation, we avert application of hand-engineered distance metrics but to learn them from data. As shown in Figure \ref{fig: pipeline}, three metric learning components ${{\varphi }_{A}}(\cdot)$, ${{\varphi }_{M}}(\cdot)$, and ${{\varphi }_{S}}(\cdot)$ are distributedly implemented in the affinity learning module. In particular, ${{\varphi }_{A}}(\cdot)$ (${{\varphi }_{M}}(\cdot)$) is formed with a sequence of fully connected layers interleaved with non-linearity, which first concatenate a pair of appearance (motion) feature vectors into a $2D_{A}$ ($2D_{M}$) dimension vector, then transform them into a scalar value indicating the pairwise affinity score. The two-stream appearance and motion cues are integrated by ${{\varphi }_{S}}(\cdot)$, which mimics a weighted summation of $a_{ij}$ and $m_{ij}$, then outputs $s_{ij}$ as final affinity score. This distributed computation of affinities renders $a_{ij}$, $m_{ij}$, and $s_{ij}$ as intermediate outputs, upon which assembled supervision can be realized. Details can be found in Section \ref{sec:41}. Affinity matrices $A$, $M$, and $S$ are generated by packing $a_{ij}$, $m_{ij}$ and $s_{ij}$ into matrix formation, and $S$ is fed into the following GNN optimization module. Additionally, the encoded appearance $F^{A}$ and motion feature vectors $F^{M}$ for each trajectory and observation are concatenated into one unified feature vector $F \in \mathbb{R} {^{{{D}_{M}} + {{D}_{A} = D}}}$. Vectors for trajectories and observations are packed together in $F_{M}$ and $F_{N}$, which are also input into the GNN optimization module.

\subsection{GNN Optimization Module}
The main obstacle of applying deep neural networks to solve data association is the varying cardinality. To cope with such variations, tedious heuristics must be designed for the adoption of deep networks who involve fixed size matrix multiplications such as MLP or LSTM.

\begin{figure}[ht]
\centering
\includegraphics[scale=0.19]{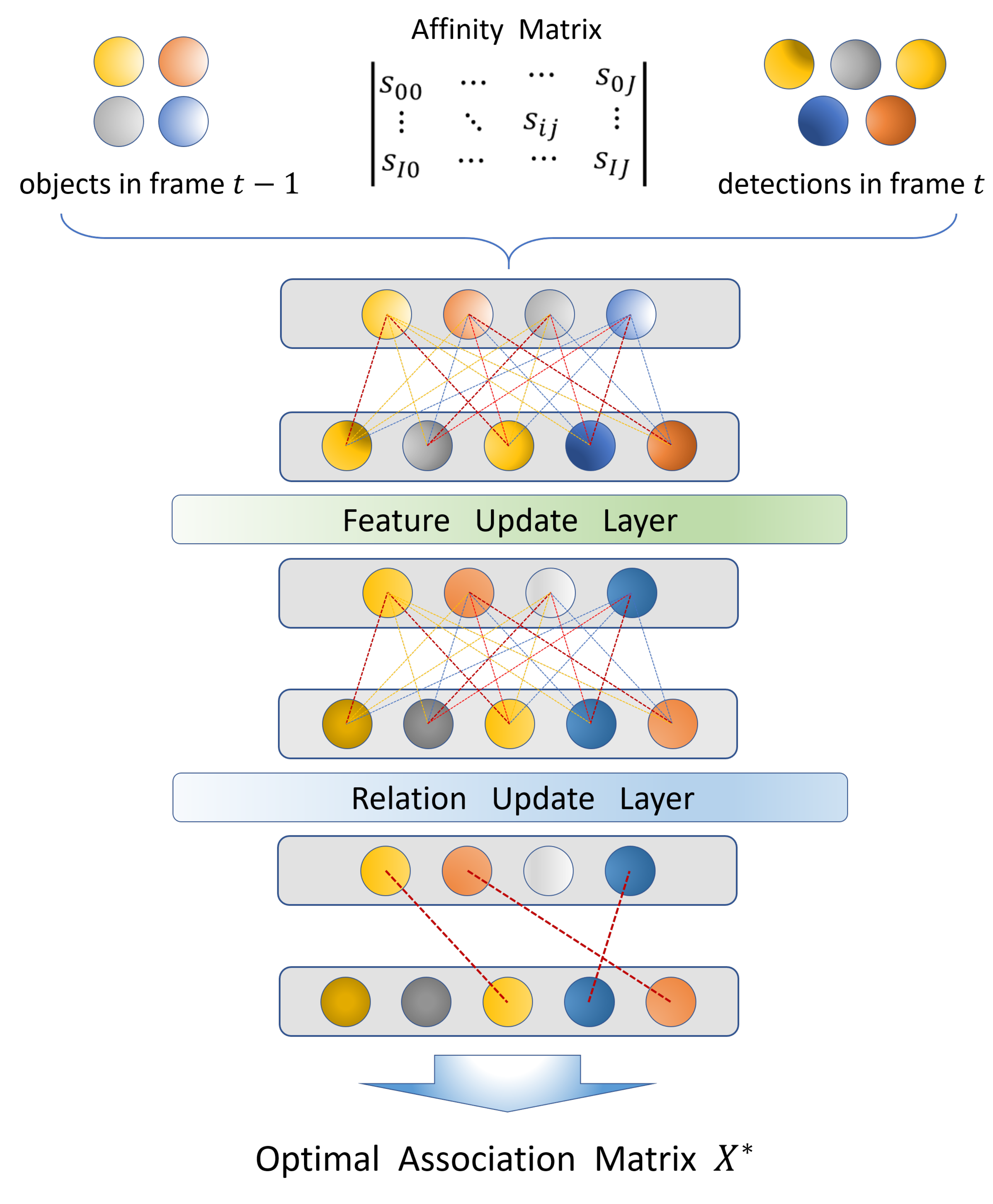}
\caption{The pipeline of the proposed optimization module based on GNN. Given the affinity matrix and the features of the objects aside in two frames, the module firstly updates their features in a message-passing way, then output their relation to output the optimal association matrix.}
\label{fig: GNN}
\end{figure}

Motivated by the graphical nature of the optimization problem, we are thus inspired to overcome the varying cardinality problem by utilizing Graph Neural Network. In comparison with LSTM \cite{Milan2017AAAI}, we believe Graph Neural Network (GNN) is more suitable in solving the maximum weighted bipartite matching problem for three reasons. Firstly, GNN operates on graphical data, therefore matches the graphical formulation of the optimization problem. Secondly, GNN deploys local operations in a message-passing way, thus can cope with the varying cardinality. Thirdly, GNN supports light-weight implementations, such that the requirement of training data is less intense. Following these motivations, a particular GNN architecture is established as shown in Figure \ref{fig: GNN}. Given the computed affinity matrix $S$ as edge weights and feature vectors $F_{M}$, $F_{N}$ as node features on a weighted in-complete bipartite $G_{t}$, the proposed GNN module composes the feature update layer and the relation update layer, who is expected to solve the optimization problem by outputting an optimal association matrix $X^{*}$, which denotes an optimal set of one-to-one associations together with accurate birth \& death indications.

\subsubsection{Feature update layer} Taking the input edge weights and node features, the feature update layer instantiates the message-passing functionality via matrix multiplications in the context of bipartite graphs, i.e. updates the feature vector for each node in one set of the graph according to all nodes in the other set weighted by the affinities in between. Considering the fact that each node in a bipartite graph only has one-hop neighbor, thus a pair of feature update layers defined as follows only need to be deployed once in the GNN to realize feature updates globally. After the message-passing step, the resulted features are further embedded into a higher dimension for enlarged model capacity.
\begin{equation}
F_M\ {\rm{ = }}\ \rho (Softmax(S)F_{N}{W_{\theta}})
\label{equa:opt5}
\end{equation}
\begin{equation}
F_N\ {\rm{ = }}\ \rho (Softmax(S^{T})F_{M}{W_{\theta}})
\label{equa:opt6}
\end{equation}
On the left side of the above equations, $F_{M} \in \mathbb{R} {^{I \times C}}$ and $F_{N} \in \mathbb{R} {^{J \times C}}$ denote the resulted features for each trajectory in the previous frame and each observation in the current frame. On the right side of the equations, $S \in \mathbb{R} {^{I \times J}}$ represents the affinity matrix computed as discussed in Section \ref{aff}. $Softmax(S)$ indicates applying softmax normalization row-wise in the affinity matrix ((\ref{equa:opt6}) in the transpose of $S$) computed by the affinity learning module. $W \in \mathbb{R} {^{D \times C}}$ indicates a set of learnable weights and $\theta$ denote the parameterizations. $\rho(\cdot)$ is the element-wise non-linearity which we adopt ReLU in this paper.

\subsubsection{Relation update layer} The updated feature vectors are fed into the relation update layer, wherein elements $x_{ij} \in \mathbb{R}$ in the association matrix $X$ are iteratively estimated by first aggregating features from a pair of nodes into the feature on the edge connecting the two, then apply a learnable transformation to compute the scalar value output. This layer can be formalized as follows:
\begin{equation}
x_{ij} = MLP_{\theta}(\sigma(F_i, F_j))
\label{equa:opt7}
\end{equation}
Here $\sigma(\cdot)$ represents the feature aggregation functionality that aggregates node features into the edge features in between. $\sigma(\cdot)$ can take many forms, in the scope of this work we realize it with non-parameterized element-wise subtraction. Basing on the aggregated edge feature, a Multilayer Perceptron parameterized by $\theta$ is employed to instantiate the transformation to get the scalar value $x_{ij}$.

\subsection{End-to-end Supervision}
It is non-trivial to supervise the training of the proposed end-to-end data association module for two reasons. Firstly, the data association result is given in $Y = X^{*}$, which is a matrix with varying dimensionality from frame-to-frame. This matrix contains both one-to-one as well as birth \& death associations, such that different supervision need to be imposed on both rows and columns. Secondly, the end-to-end framework establishes a network composing components for various tasks, training needs to be carefully designed to facilitate back-propagation and to avoid gradient vanishing problem. To overcome these difficulties, we first clarify the ground truth matrix generation, then propose the multi-level matrix loss in conjunction with assembled supervision.

\subsubsection{Ground Truth Generation}
To realize matrix-wise supervision, we generate ground truth association matrix $\hat{Y}$ with elements $\hat{y_{ij} \in {0,1}}$. As defined in (\ref{equa:opt2}), there exists a sub-matrix $Y_{O2O} \in \mathbb{R}^{k \times k}$ in $Y$ that conforms to one-to-one associations. Corresponding rows and columns in $\hat{Y}$ are generated as one-hot vectors with $1$ placed at the $y_{ij}$, indicating row (trajectory) $i$ occupies detection (column) $j$. $Y_{B\&D}$ (${{Y}_{B\&D}}\cup Y_{O2O}=Y$) indicates birth \& death associations, corresponding row and column vectors in $\hat{Y}$ are generated with all $0$ elements.

\subsubsection{Multi-level Matrix Loss}
The loss computed on an estimated association matrix $Y \in \mathbb{R} {^{{m} \times {n}}}$ and its corresponding ground truth $\hat{Y}$ is defined as a combination of element-level and vector-level losses. On the element-level, each element in the matrix is the estimation for a binary classification, specifying whether this element denoting a match or mismatch. Accordingly, a binary cross-entropy loss $L_{e}$ formulated as follows is applied to each element to supervise the classification:
\begin{equation}
\begin{split}
{L_{e}} = \sum\limits_{i}^{I}{\sum\limits_{j}^{J}({- {p}{\hat{y_{ij}}}\log \sigma ({y_{ij}}) - (1 - {\hat{y_{ij}}})\log (1 - \sigma ({y_{ij}}))}}),
\label{equ:elementLoss}
\end{split}
\end{equation}
where ${y_{ij}} \in Y, \hat{y_{ij}} \in \hat{Y}$. $p$ is the weighting factor assigned to positive examples to alleviate the sample imbalance. In our experiments $p$ equals to 25.

On the vector-level, we separate the one-to-one association from the birth \& death, and denote the sub-matrix as $Y_{O2O}$ and $Y_{B\&D}$ respecitively. For vectors $v_{O2O}$ within $Y_{O2O}$, a cross entropy loss $L_{O2O}$ is adopted to supervise a multi-class classification composing the estimated vectors and the one-hot ground truths:
\begin{equation}
{{L}_{O2O}}=-\sum\limits_{O2O}^{k}{\hat{v}_{O2O}}\log (softmax(v_{O2O})),
\label{equ:Vloss1}
\end{equation}
where $\hat{v}_{O2O}$ denotes corresponding one-hot ground truth vector, $k$ denotes the number of associations.

For vectors in $Y_{B\&D}$, we apply a mean square error (MSE) loss $L_{B\&D}$ to enforce the vector to approach to negative infinity, which can be easily recognized in the tracking process:
\begin{equation}
{{L}_{B\&D}}={{\sum\limits_{B\&D}^{v}}{\left\| {sigmoid({v}_{B\&D})} \right\|}^{2}},
\label{equ:Vloss2}
\end{equation}
where $v = m + n -2*k$.

The multi-level matrix loss $L_{matrix}$ is then computed as a summation of losses on different levels over the matrix:
\begin{equation}
{{L}_{Matrix}}={{L}_{E}}+{{L}_{O2O}}+{{L}_{B\&D}}
\label{equ:VlossCombo}
\end{equation}

\subsubsection{Assembled Supervision}
Instead of only computing $L_{e}$ on the final output matrix $Y$, we also assemble it computed on affinity scores $A$, $M$, and $S$. During back-propagation, the gradient flows start at each matrix and flow backwards distributedly. As such, the gradients on earlier network layers are a summation of each flow, therefore the gradient is enhanced and the vanishing problem is alleviated.

\subsection{Association Result Interpretation}
The optimal association matrix $X^*$ produced by the optimization module contains indications for both one-to-one and object death \& birth associations. The elements in $X^*$ is not binary indicators directly but can be easily interpreted. Conforming to our training setup as detailed in Section \ref{sec:41}, each element is the result of a binary classification, where the one-to-one association is marked as positive. As a result, the one-to-one association is indicated by the largest positive value in a row, and the trajectory corresponding to this row occupies the detection denoted by the corresponding column that largest value resides in. Each one-to-one association is denoted with an indicator in $O\in {\mathbb{R}^{2}}$. Birth or death happens when a row or column contains only negative values, and each of them is marked with an indicator in $B\in {\mathbb{R}}$ and $D\in {\mathbb{R}}$. To efficiently integrate $O$, $B$, and $D$ from $X^{*}$, we follow a straightforward procedure by iteratively finding the largest element $x_{max} = x_{ij}^{*}$ in $X^{*}$. If $x_{max} > 0$, then add ${(i,j)}$ into $O$, and row $i$ as well as column $j$ is marked unavailable. Once added $x_{max}$ is negative, then all the available but un-associated rows and columns are added into $D$ and $B$. Accordingly, the final association result is given indicators in $O$, $B$, and $D$.

\section{Implementation}
\label{sec:4}
The implementation of our method is separated into two parts: the training methodology and the tracking methodology.

\subsection{Training Methodology}
\label{sec:41}
The proposed model is implemented as follow. For the network design, we construct the CNN in the Siamese network with only four convolutional layers inserted with the ReLU nonlinearity. For the LSTM module, we set the length of tracklet $L = 5$. We use the ground-truth bounding boxes and target IDs in the MOT17 \cite {mot16} training set to generate pairs of images and motion information. All crop images are resized to ${84 \times 32}$ to maintain the aspect ratio of targets. All modules are trained from scratch. The proposed model is trained end-to-end with Adam \cite{adam} optimizer, where weight decay is set to $0.0005$ and initial learning rate is set to $lr = 0.001$, divided by $10$ for every $10000$ iterations, and $40000$ iterations in total.
\begin{figure}[ht]
\centering\
\includegraphics[scale=0.40]{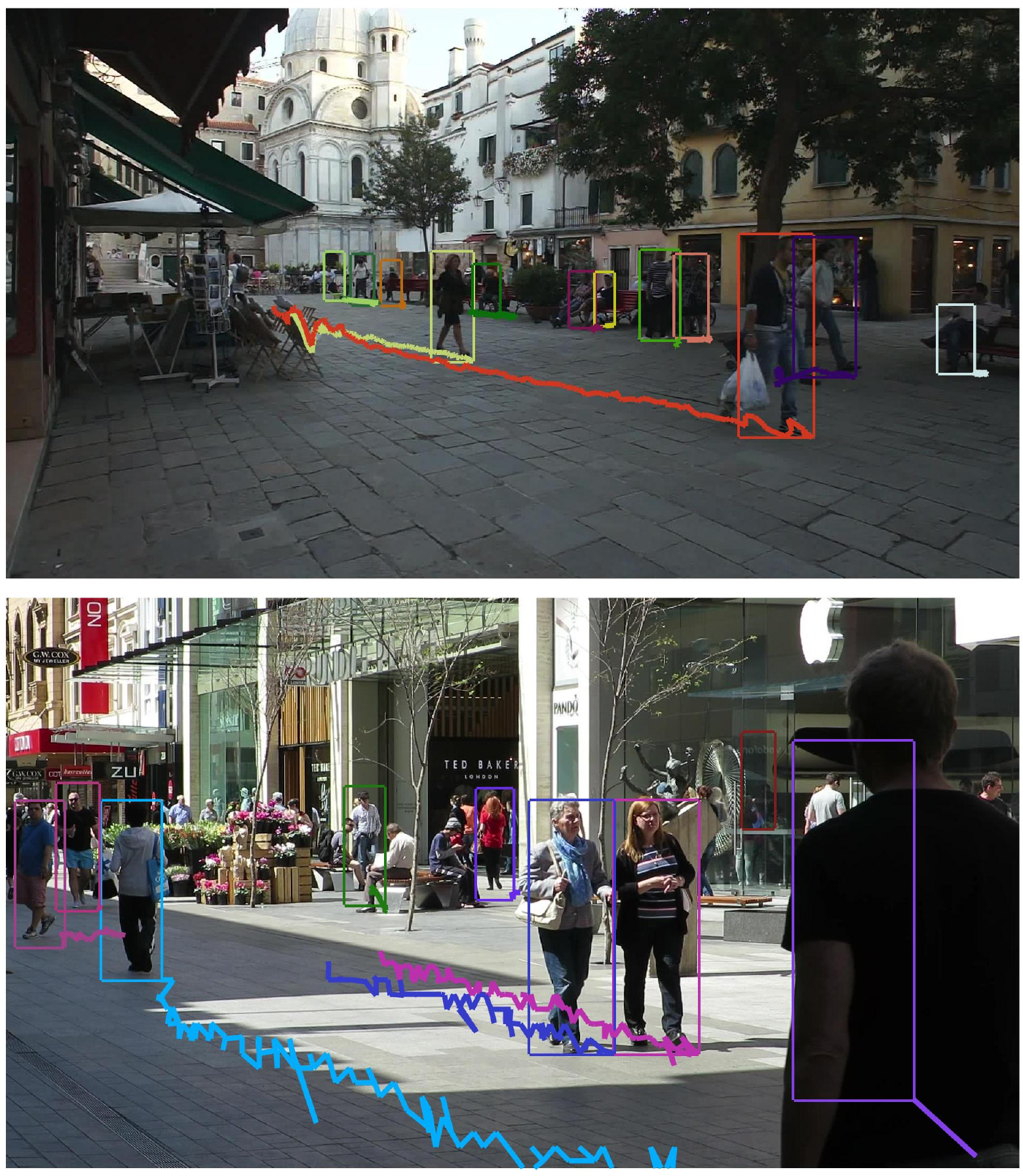}
\caption{Visualization of tracking results on MOT17 benchmark dataset. Displayed frames are from the testing sequence 01 and 08.}
\label{fig: demo}
\end{figure}

\subsection{Tracking Methodology}
\label{sec:42}
To limit false births \& deaths induced by noisy detections, we set up a time window with length $T_b$ to verify an object birth. Specifically, once a birth is indicated in $B$, the tracker postpones the trajectory initialization process until this object consecutively appears in $O$ for the next $T_b$ frames. Similarly, $T_d$ is set up for confirming the death. Once $D$ reports a death, we deploy a dummy observation with the object's last seen bounding box dimension but propagated into future frames with a linear motion model calculated using the trajectory's short tracklet of coordinates. If the dummy fails to be associated in $T_{d}$ frames, it is terminated. We set $T_b = fps / 2$ and $T_d = fps / 6$ ($fps$ denotes the frame rate), according to the model's performance on the training set.

\section{Experimental Results}
In this section, we present our experimental results on 2D MOT 2015 \cite{mot15} and MOT17 \cite{mot16} benchmark datasets with ablation studies and comparisons with selected baselines. More details for the MOT benchmark are available at \textit{https://motchallenge.net}.

\begin{table*}
\caption{Tracking Performance on the MOT17 Benchmark}
\begin{center}
\resizebox{175mm}{20mm}{
\begin{tabular}{|c|c|c|c|c|c|c|c|c|c|c|}
\hline
Mode & Tracker & MOTA$\uparrow$  & MOTP$\uparrow$ & IDF1$\uparrow$ & ID Sw.$\downarrow$  & MT$\uparrow$ & ML$\downarrow$  & Frag$\downarrow$  & FP$\downarrow$  & FN$\downarrow$  \\
\hline\hline
Online & GM\underline{ }PHD \cite{GM_PHD} & 36.4 & 76.2 & 33.9 & 4,607 & 4.1$\%$ & 57.3$\%$ & 11,317 & \textbf{23,723} & 330,767 \\\hline
Online & GMPHD\underline{ }KCF \cite{GMPHD_KCF} & 39.6 & 74.5 & 36.6 & 5,811 & 8.8$\%$ & 43.3$\%$ & 7,414 & 50,903 & 284,228 \\\hline
Online & EAMTT \cite{EAMTT} & 42.6 & 76.0 & \textbf{41.8} & 4,488 & 12.7$\%$ & 42.7$\%$ & 5,720 & 30,711 & 288,474 \\\hline
Online & \textbf{Ours} & \textbf{45.5} & \textbf{76.3} & 40.5 & \textbf{4,091} & \textbf{15.6$\%$} & \textbf{40.6$\%$} & \textbf{5,579} & 25,685 & \textbf{277,663} \\\hline
\hline
\hline
Offline & MHT\underline{ }bLSTM \cite{Eccv18BilinaerLSTM} & 47.5 & 77.5 & 51.9 & 2,069 & 18.2$\%$ & 41.7$\%$ & 3,124 & 25,981 & 268,042 \\\hline
Offline & IOU17 \cite{IOU17} & 45.5 & 76.9 & 39.4 & 5,988 & 15.7$\%$ & 40.5$\%$ & 7,404 & 19,993 & 281,643 \\\hline
Offline & EDMT17 \cite{EDMT17} & 50.0 & 77.3 & 51.3 & 2,264 & 21.6$\%$ & 36.3$\%$ & 3,260 & 32,279 & 247,297 \\\hline
\end{tabular}}
\end{center}
\label{tab: MOT17}
\end{table*}

\begin{table*}
\caption{Tracking Performance on the MOT15 Benchmark}
\begin{center}
\resizebox{175mm}{10mm}{
\begin{tabular}{|c|c|c|c|c|c|c|c|c|c|c|}
\hline
Mode & Tracker & MOTA$\uparrow$  & MOTP$\uparrow$ & IDF1$\uparrow$ & ID Sw.$\downarrow$  & MT$\uparrow$ & ML$\downarrow$  & Frag$\downarrow$  & FP$\downarrow$  & FN$\downarrow$  \\
\hline\hline
Online & RMOT \cite{RMOT} & 18.6 & 69.6 & \textbf{32.6} & \textbf{684} & 5.3$\%$ & 53.3$\%$ & \textbf{1,282} & 12,473 & 36,835 \\\hline
Online & RNN\underline{ }LSTM \cite{Milan2017AAAI} & 19.0 & \textbf{71.0} & 17.1 & 1490 & 5.5$\%$ & 45.6$\%$ & 2,081 & \textbf{11,578} & 36,706 \\\hline
Online & \textbf{Ours} & \textbf{21.8} & 70.5 & 27.8 & 1488 & \textbf{9$\%$} & \textbf{40.2$\%$} & 1,851 & 11,970 & \textbf{34,587} \\\hline
\end{tabular}}
\end{center}
\label{tab: MOT15}
\end{table*}

\begin{table}
\caption{Ablation Study on the MOT17 Benchmark}
\begin{center}
\resizebox{82mm}{10mm}{
\begin{tabular}{|c|c|c|c|}
\hline
Configurations & MOTA$\uparrow$  & ID Sw.$\downarrow$  & MT$\uparrow$ \\
\hline\hline
w/o assembled supervision & 43.2 & 4275 & 14.4\% \\\hline
w/o optimization module & 38.6 & 9197 & 11.8\% \\\hline
\textbf{Full} & 45.4 & 4126 & 15.6\% \\\hline
\end{tabular}}
\end{center}
\label{tab: Ablation1}
\end{table}

\subsection{Evaluation Metrics}
\label{sec: 51}
The evaluation metrics used on MOT benchmarks include Multiple Object Tracking Accuracy (MOTA), Multiple Object Tracking Precision (MOTP), ID F1 score (IDF1), ID Precision (IDP), ID Recall (IDR), Mostly Tracked trajectories (MT, the ratio of ground-truth trajectories that are at least 80\% covered by the tracking output), Mostly Lost trajectories (ML, the ratio of ground-truth trajectories that are at most 20\% covered by the tracking output), number of False Negatives (FN), number of False Positives (FP), number of ID Switches (ID Sw.), number of Track Fragmentations (Frag).

\subsection{Evaluation on MOT Benchmark}
\label{sec: 52}
The evaluation results on both MOT17 and MOT15 datasets are shown in Table \ref{tab: MOT17} and \ref{tab: MOT15}. The arrows in each column denote the favorable changing direction of the corresponding metric. Being the only fully end-to-end trained online tracker, we emphasize highlighting the benefits of the end-to-end training methodology as well as the GNN optimization module. To this end, we avoid excessive parameter tuning efforts during testing, and training data augmentation, as well as post-tracking performance boosting heuristics refrain in the experiments. Nevertheless, we still demonstrate competitive results with other published online trackers in both datasets. Particularly, on MOT15, we compare our results with the RNN\underline{ }LSTM \cite{Milan2017AAAI}, which is claimed the first fully end-to-end multi-object tracking method that inspires our work. As shown, we achieve favorable results on several important evaluation metrics, including 39.5\%, 12.8\%, and 3.5\% improvements on IDF1, MOTA, and MT. These improvements results from the fact we integrate the appearance affinity into the end-to-end framework and the GNN optimization module can better cope with the varying cardinality problem than RNN and LSTM.

\subsection{Ablation Study}
\label{sec: 53}
Ablation study is also conducted on the MOT17 benchmark. As shown in Table \ref{tab: Ablation1}, we demonstrate the contributions of the assembled supervision and the GNN optimization module. Specifically, the first row of the table shows the performance of the whole network trained with single supervision applied on the final association output $Y$ without assembled supervision. The second row illustrates the result of directly reasoning the association result using the affinity matrix $S$, without the following GNN optimization module. Training for this configuration is assembled on affinity matrices $A$, $M$, and $S$. The last row reports the full network. As illustrated, the full network outperforms the first-row configuration in all three metrics with 4.9\%, 3.5\%, and 7.7\% improvements, proving the merits of the assembled supervision. Comparing to the second-row configuration, the full network enlarges the improvements to 15\%, 122.9\%, and 24.4\%. Although this configuration cannot be trained with the full assembled supervision due to the absence of $Y$, the extra performance improvements still advocate the contribution of the GNN optimization module.

\section{Conclusion and Future work}
We propose an end-to-end data association model for online multi-object tracking. By jointly training the affinity learning module and the GNN optimization module, they can co-adapt collaboratively, improving the adaptivity, scalability, and accuracy of the data association model. Particular, we firstly introduce GNN in the context of solving online data association with frequent birth and death, successfully settles the irregular linear assignment formulation in a data-driven way. In this paper, we emphasize on demonstrating the efficacy of end-to-end data association and the GNN optimization module, therefore the affinity computation module is lightly designed and no performance enhancing heuristics have been employed. The performance of the method can be further enhanced in future work.

{\small
\bibliographystyle{ieee}
\bibliography{egbib}
}

\end{document}